# AUTOGENIC TRAINING WITH NATURAL LANGUAGE PROCESSING MODULES: A RECENT TOOL FOR CERTAIN NEURO COGNITIVE STUDIES

S. Ravichandran * and M. N. Karthik #
* Biomedical Informatics and Engineering division
Temasek Engineering School, Temasek Polytechnic, Singapore.
# National Institute of Technology & Science, University of Madras, India

## Introduction

Learning to respond to voice-text inputs involves the subject's ability in understanding the phonetic and text based contents and his/her ability to communicate based on his/her experience. The neuro-cognitive faculty of the subject has to support two important domains in order to make the learning process complete. In many cases, though the understanding is complete, the response is partial. This is one valid reason why we need to support the information from the subject with scalable techniques such as Natural Language Processing (NLP) for abstraction of the contents from the output. This paper explores the feasibility of using NLP modules interlaced with Neural Networks to perform the required task in autogenic training related to medical applications.

## Methods

In any Knowledge Based System, knowledge engineering does not end with the collection of the bulk of the information for the knowledge base. The key task involved in interactive knowledge abstraction depends on how well inference engines can extract the relevant data from the bulk of the information available in the knowledge base [1]. This is perhaps the reason why we need interactive tools and powerful interfaces for knowledge abstraction and knowledge translation.

Autogenic training as a cognitive tool in behaviour abstraction is a well-recognized method in medical practice and is one the most sought after tools in clinical practice [2]. A simple autogenic training system would consist of few input devices to receive information and an output device to assist the subject in the rehabilitative process. The drawbacks in such a simple system is that it does not work on a complete knowledge domain assisted by the required inference engines. Moreover the major deficiency of a simple system is that it is incapable of adapting its framework for new challenging problems.

## Key issues in autogenic training

Though there are many issues, which are of equal importance in performing an autogenic training programme, we restrict ourselves to a few key issues, which are very important in conducting an autogenic programme with well-defined cognitive strategies. The first one is focussed on how a given input to the system is received and processed. The second issue is on the use of phonetic and text based NLP for data abstraction. The last one will be on how the NLP system can be used to train the neural networks for case based guided response.

## Architecture and supporting modules of the system

The speech input from the subject is digitised and converted into an NLP Parser understandable expression. This is evaluated based on existing inputs stored in a knowledge base using a request/response data translator. An inference engine interacts between the knowledge base and the output driver to arrive at the right inference using case based reasoning. The output driver converts these inferences into subject understandable outputs.

Besides the NLP Parser and the translator, there are three other modules, which form the integrated part of the system. These are the speech digitiser, inference engine and the output driver. The speech digitiser used in this framework is of hybrid type, which converts speech into digital data. This data is then converted into phonemes, which are compatible with the NLP Parser. The inference engine of the system does a case based search in the knowledge base, depending on the response of the Boltzmann network based translator. Based on the Pragmatic knowledge, the inference engine then converts the result into context specific outputs. These outputs are fed into the output driver The output driver converts the result obtained into user understandable output. This output is usually an audio output, but there are provisions to provide a visual/text output.

## NLP in data abstraction

NLP is one of the methodologies involved in evaluating the data provided as an input to a system in terms of logical linguistic expressions. The gathering of data for NLP involves different levels of language analysis, like phonetic & phonological knowledge, morphological knowledge, syntactic knowledge, semantic knowledge, pragmatic knowledge and discourse knowledge. Ideally, all the above knowledge domains are essential to the functioning of a full-fledged NLP system [2]. Since it is beyond the scope of this paper to deal in-depth with all the facets of NLP, we shall restrict ourselves to the relevant domains, which are essential in our medical application on autogenic training.

In the present autogenic trainer, we would concentrate on certain essential data, which will be gathered from the user. We shall consider a simple speech recognition system that will convert every sound into its constituent phonemes. These phonemes will then be converted into known words wherever possible, or left as such if there is no available information.

The conversions of the phonemes to the words are performed using the N-gram technique [3]. In our application, this conversion uses a mix of N-gram techniques with phonemes and morphemes, to arrive at the best possible match.

**Interfacing with neural networks**

The heart of the autogenic trainer consists of two sets of neural networks. These networks are trained to extrapolate from minimal etymological input. Enacting morphological actions on the etymological inputs performs the extrapolation, till a best possible semantically correct match is reached. The first neural network is a Hopfield network, which works from within the NLP parser. This is based on a Lexical database, and hence there is a need for the neural network to settle into local minima. This is the reason why a Hopfield network has been chosen.

The second neural network, which is built into the translator, is a Boltzmann machine, which tries to, learns & analyse various request/responses and arrive at the best possible match based on earlier inputs. This consists of a Bayesian analyser to correlate various responses. It also calculates the various relationships between the data and extrapolates the established patterns. The neural net is built in such a way that from the inputs of the various states that it receives, it would be able to retrace itself, as well as learn to identify new patterns that emerge as newer action. Hence, there is a need for a neural network which can settle into a globally optimal state satisfying as many interacting constraints and solutions as possible. Thus, the use of a Boltzmann network for the request/response translator system is justified [4].

The nodes of the neural networks would be built using weightage based fuzzy (logic) variables. This way, alternate patterns could be mapped into the network, so that the system is capable of arriving at more than one solution. These new mappings could be created dynamically which would again scale the engine into predicting more than one action for a given pattern. All new patterns are analysed statistically, based on a number of parameters, hence the system can adapt itself to a variety of changes.

**Training the neural networks**

We have considered three possible neural networks configurations, namely the Hopfield network, Perceptron network and the Boltzmann machines . For the training, N-gram technique was applied on the morphological and phonological constituents of the word(s) corresponding to the etymological origin. The obtained morphemes and phonemes were used as the deterministic parameters of reference for inputs.

The neural networks keep all data that are more likely to occur in future inputs. There is a pruning of unnecessary parameters based on the N-gram distribution. This way a probabilistic modelling of the N-grams is performed, and all N-grams with the probability less than a threshold probability are removed [5].

All the three neural network configurations were tested for their precision with a standard sample case base. The following table provides preliminary statistical information on the precision of the three types of neural networks that were used in our application –

| Type of Network | Precision at NLP stage | Precision at I/O stage | Adaptive Capability |
|---|---|---|---|
| Hopfield | 66% | 54% | Very High |
| Perceptron | 54% | 55% | Medium |
| Boltzmann | 48% | 58.5% | High |

**Conclusion**

The present approach to interactive autogenic training is only a means to provide a meaningful link between the subject and a high-end computer assisted tool supported by NLP techniques.

The techniques presented in this paper are aimed at improving the interaction between the subject and the system.

From the clinical and statistical data obtained, we feel that there is still scope for improving the effectiveness of the system in terms of user interaction and enhancement of the NLP Parser for better extrapolative capabilities.

Though there is no doubt that a clinical psychologist as a trainer would be the most ideal person to improve the cognitive process of the subject, it has been observed that many computer based cognitive tools play a very valuable role in the learning process. This is especially true, in case of auditory and visual style non-kinaesthetic learners.

There is a sizeable amount of research still in store for researchers in AI; we believe that this paper is an honest attempt at integrating the various tools of engineering and research for cognitive support systems